\documentclass[letterpaper, 10 pt, conference]{ieeeconf}  

\IEEEoverridecommandlockouts                              

\usepackage{graphics} 
\usepackage{epsfig} 
\usepackage{listings}
\usepackage{subcaption}
\usepackage{algpseudocode}
\usepackage{algorithm}

\title{\LARGE \bf
Multi-Robot Strategies for Communication-Constrained Exploration and Electrostatic Anomaly Characterization
}

\author{Gjosse Zijlstra$^1$, Karen L. Aplin$^2$, Edmund R. Hunt$^1$
\thanks{$^{1}$School of Engineering Mathematics and Technology, Faculty of Engineering, University of Bristol, Bristol, UK. \newline Email:
gjosse.zijlstra@bristol.ac.uk}
\thanks{$^{2}$School of Civil, Aerospace, and Design. Faculty of Engineering, University of Bristol, Bristol, UK.}%
\thanks{GZ is supported by the UK EPSRC (EP/L015293/1). ERH is supported by the Royal Academy of Engineering under the Research Fellowship program.}
}

\begin{document}

\maketitle
\thispagestyle{empty}
\pagestyle{empty}

\begin{abstract}
Exploration of extreme or remote environments such as Mars is often recognized as an opportunity for multi-robot systems. However, this poses challenges for maintaining robust inter-robot communication without preexisting infrastructure. It may be that robots can only share information when they are physically in close proximity with each other. At the same time, atmospheric phenomena such as dust devils are poorly understood and characterization of their electrostatic properties is of scientific interest.  We perform a comparative analysis of two multi-robot communication strategies: a distributed approach, with pairwise intermittent rendezvous, and a centralized, fixed base station approach. We also introduce and evaluate the effectiveness of an algorithm designed to predict the location and strength of electrostatic anomalies, assuming robot proximity. Using an agent-based simulation, we assess the performance of these strategies in a 2D grid cell representation of a Martian environment. Results indicate that a decentralized rendezvous system consistently outperforms a fixed base station system in terms of exploration speed and in reducing the risk of data loss. We also find that inter-robot data sharing improves performance when trying to predict the location and strength of an electrostatic anomaly. These findings indicate the importance of appropriate communication strategies for efficient multi-robot science missions.

\end{abstract}
\section{Introduction}
Exploring extreme environments with multi-robot systems, such as Mars or the Moon, presents many challenges. One such challenge is having reliable and effective communication links between robots to allow data transfers, especially if the robots travel far away from each other and there is no supporting communication infrastructure. Meanwhile, a scientific opportunity in exploring other planets is the lack of research into phenomena in atmospheric electricity, an example of which are dust devils, which are suspected to be a source of Martian lightning \cite{ruf2009emission}. Here, we compare two multi-robot communication strategies: a distributed approach with intermittent rendezvous between pairs of robots, and a fixed base station approach. In both strategies, exploration data can only be shared within a certain proximity. We also consider how we can use multi-robot systems to locate a dust devil and estimate both its position and strength, assuming communications proximity in the first instance. 

Our goals are to first, determine which communication strategies can explore the environment the fastest and also evaluate the amount of data at risk if the system loses a robot. This research can contribute to decisions about which communication architectures are useful when multi-robot systems are considered for exploration of Mars, the Moon, or communication-constrained environments on Earth. We hypothesize that the intermittent pairwise rendezvous communication system will demonstrate a faster exploration rate compared to the centralized base approach, and show a lower risk of data loss because of more frequent meet-ups. This is based on the notion that the flexibility of intermittent rendezvous might allow for more dynamic data sharing and shorter navigational paths for agents to travel. Our second goal is to consider the opportunity and potential approach to using multi-robot systems to detect and characterize electrostatic anomalies such as dust devils. We hypothesize that the availability of multiple, distributed gradient readings in a multi-robot system will allow significantly quicker localization and characterization of anomalies. We present some background on recent extraterrestrial exploration missions, electrostatic research and dust devil phenomena before describing our algorithm and simulation approach.

\section{Background}
\begin{figure}[h]
    \centering
    \includegraphics[width=0.5\columnwidth]{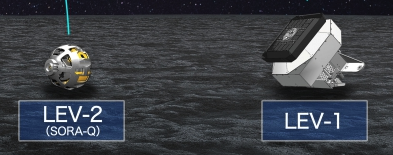}
    \caption{JAXA moon robots, LEV-1 and LEV-2.}
    \label{fig:jaxa}
\end{figure}

\begin{figure}[h]
    \centering
    \includegraphics[width=0.5\columnwidth]{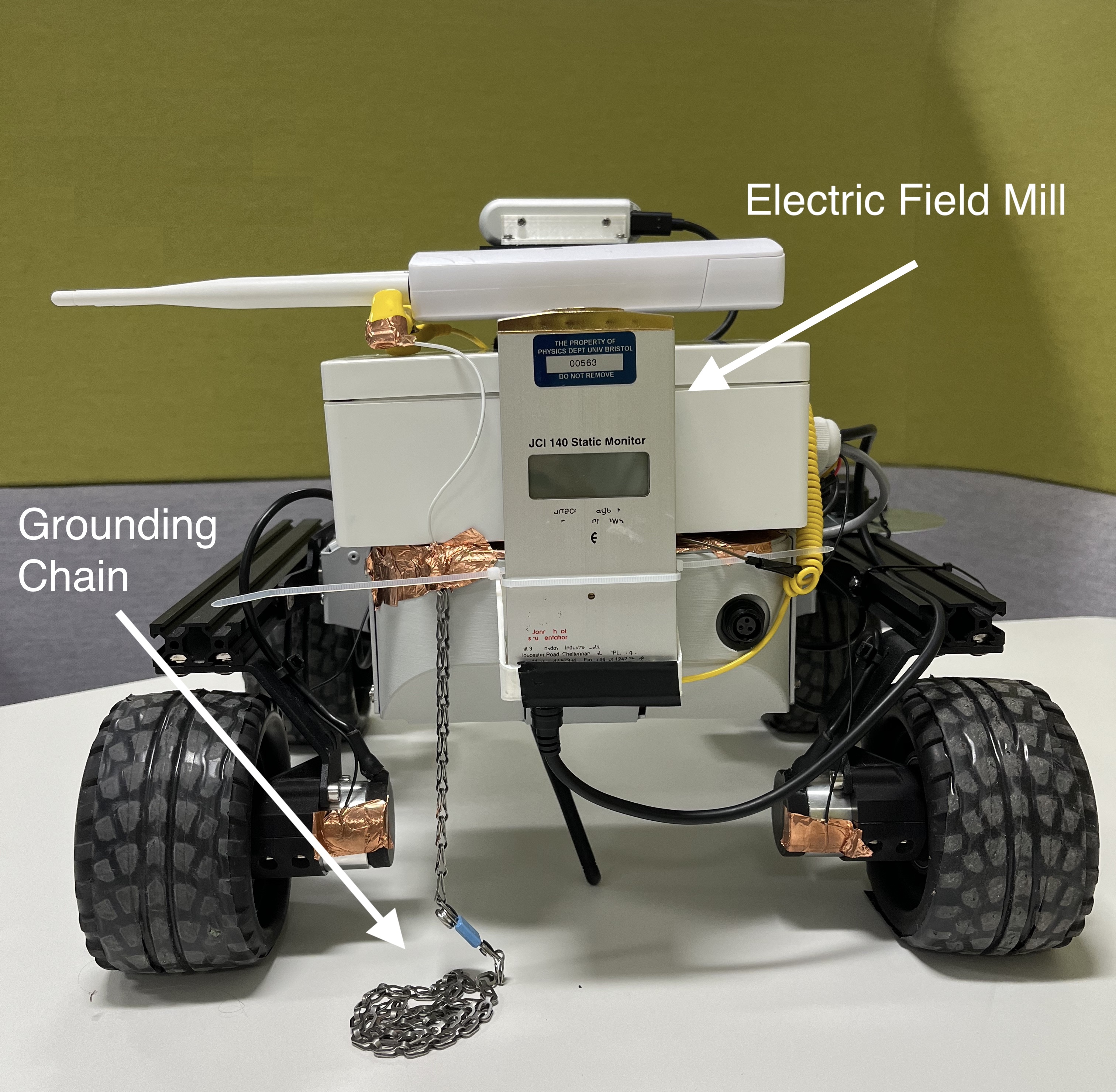}
    \caption{Rover robot with electric field mill for electrostatic electricity mapping \cite{zijlstra2024}.}
    \label{fig:rover}
\end{figure}

In robotic Mars missions to date, exploration has only been performed by one rover, although the Perseverance Rover has carried a small helicopter, Ingenuity \cite{Balaram2021}. Ingenuity served multiple purposes: among these was acting as a technology demonstrator, showcasing the first instance of multi-robot communication on Mars. The radio systems onboard Ingenuity utilize a power-saving approach, with a range around 250 m for higher bandwidth settings and around 1 km when used in lower bandwidth settings. \cite{Chahat2020helicopter}. 
There is ongoing research into the potential to use multi-robot systems, which could comprise, for example, smaller rovers \cite{Laine2018} or legged robots \cite{Kolvenbach2019,Uno2021}. Currently, there is a Moon mission by JAXA (Japan Aerospace Exploration Agency) where they have deployed two heterogeneous smaller robots \cite{jaxa}, seen in Fig.~\ref{fig:jaxa}  In addition to mapping physical terrain, robots could be used for mapping other physical variables of note; we are particularly interested in atmospheric electricity as a motivating example \cite{leblanc2008planetary}. We have considered the use of electric field mills mounted to rover robots, which might realistically number 4--6 (Fig.~\ref{fig:rover}, \cite{zijlstra2024}). One motivation to measure the electrostatic field is to detect and locate dust devils. Dust devils are atmospheric vortices which can vary considerably both in width (0.5--30 m) and height (30 m--2 km, \cite{dustdevil,nasadust}). The dust particles that get picked up rub against each other (triboelectrically charged) and create large electrostatic fields, which can be measured remotely \cite{crozier1970dust}. To date, there have been no measurements of the electrostatic fields created by dust devils on Mars but there are simulation studies of dust devils on Earth and on Mars, which estimate the limits of dust devil electric discharge \cite{zhai2006quasi}. In order characterize dust devils using a multi-robot system we develop an algorithm to turn the locally read electrostatic field measurements into a prediction of a dust devil's location and strength. Dust devils are poorly understood, and analyzing the direction and velocity of them would allow for a greater understanding of the underlying physical mechanisms. An example of past research into multi-robot algorithms for environment anomaly detection include CIMAX (`Collective Information Maximization') \cite{hornischer2021cimax}, a decentralized approach that assumes a rather high density of swarm robots that may not be feasible in space missions. The use of \textit{virtual} potential fields is quite common as a control approach for robot path planning (e.g. \cite{Valavanis2000,Pimenta2006,Bayat2018}), and has been considered in the context of planetary exploration \cite{sancho}, although navigation informed by the physical electrostatic field has not, to our knowledge, been considered.

Key elements of deploying a multi-robot system are effective exploration and communication strategies. An example of a real-time Coverage Path Planning (CPP) solution is presented in \cite{Laine2021}. With regard to communication, there is research into maintaining connectivity with the use of micro-robots on lunar exploration missions \cite{Paet2021}. Paet et al. \cite{Paet2021} describe a way of maintaining a connection between robots by using radio propagation and communication graphs. They simulated a network of three robots whereby the rovers' mobility affected the network connectivity. A multi-robot deployment to a remote and/or extreme environment such as Mars may have robots necessarily far apart and hence unable to communicate high volumes of data easily until they are in proximity again. In this case, a decentralized communication architecture could be advantageous, because long journeys will be necessary back and forth to a central hub robot or `mothership'. A `sparse swarm' configuration \cite{Tarapore2020} could be the logical choice in that case; in exploration, it could employ some form of `Swarm SLAM' \cite{Kegeleirs2021}. This would entail each robot building its own map of the environment and then periodically sharing it with other robots; yet appropriate, distributed data sharing strategies are still in development for swarm SLAM, both among robots and relaying back to human users \cite{Kegeleirs2021}. We consider a data-sharing strategy where robots intermittently rendezvous to share their data, to safeguard and enlarge the map they have built. 

\section{Methods}
We present the Methods in two parts, \textit{A} and \textit{B}: for the communication-constrained exploration strategies and electrostatic anomaly characterization approaches, respectively. We use a simulator based on the Python \textit{Mesa} library \cite{masad2015mesa}, an agent-based modelling (ABM) framework with a grid-based environment. Our custom-designed agents (robots) and models are integrated into Mesa. This allows for flexibility in testing different scenarios and algorithms. 

\subsection{Communication-Constrained Exploration}
In these experiments, we evaluate two approaches on two different maps, one with inaccessible terrain features and another with flat, featureless terrain. Simulations will be carried by a number of robots (from 2 to 5, representing a realistic mission payload) and different values for the `$\mathrm{connectionTime}$' threshold (see Algo. 1 later). These trials aim to assess the speed of environment exploration and to monitor the amount of unique data in the system, which indicates potential data loss if a robot is lost. Each configuration is simulated 20 times. 

\subsubsection{Environment Description}
\begin{figure}[h]
    \centering
    \includegraphics[width=0.2\textwidth]{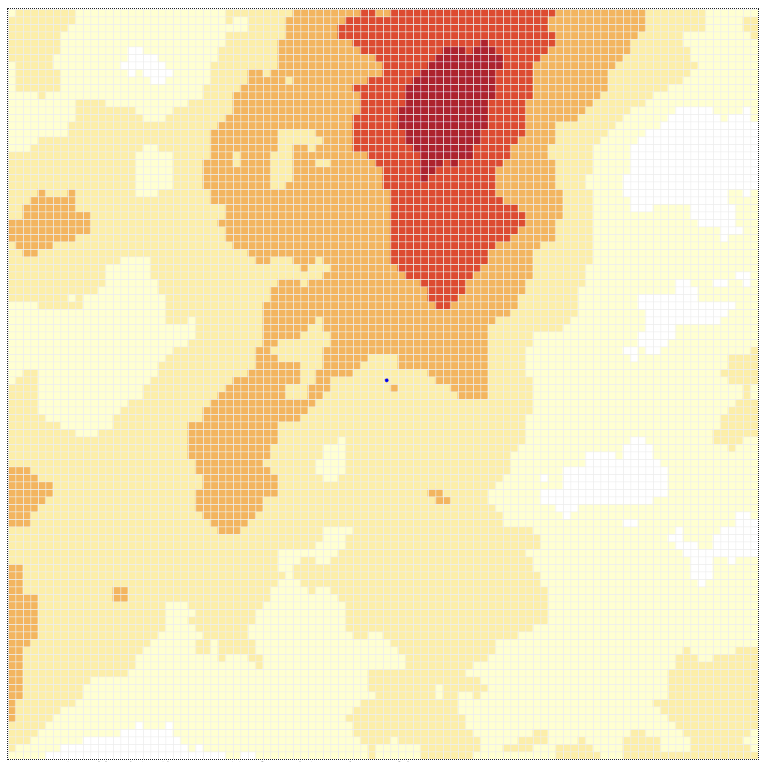}
    \caption{Simulated Martian terrain map with rugged and impassable features}
    \label{fig:marsworld}
\end{figure}

A Mars map is generated using random noise to create a 100$\times$100 cell world with a hazard level, that ranges from 1--5, whereby 5 is the maximum hazard level. One of the worlds used in the simulations can be seen in Figure~\ref{fig:marsworld}, where more red cells represent a higher hazard level. The other baseline map is the same 100$\times$100 however there are no hazard levels defined. The scale of the environment is considered to be 10 km $\times$ 10 km to have a realistic limitation of communication range. 

For each map, two different communication systems are tested. These have the same exploration strategy for robots based on a greedy approach. The algorithm works as follows: at each step, a robot will look at its known map of cells, find any unknown cells bordering those, choose a random one to move to, and generate a path using the A* algorithm. While the robot is moving it has a radius around it that will ``see" cells. Every time a new cell is ``seen" the robot gains data, which is defined by: 
\begin{lstlisting}[language=Python] 
self.dataLevel += 1 + random.randint(0, 3)
\end{lstlisting}
This algorithm is described in more detail in Algorithm 1. 

\subsubsection{Fixed Base Station Communication System}
The first system that will be tested and used as a baseline is a fixed base station, or centralized system. In this simulation, there is a fixed robot at the center of the map that will not move. Once a robot decides to transfer data, the robot will travel with the most efficient route back to the fixed point and have a bi-directional transfer of its data. To determine when to go back and transfer a variable called `$\mathrm{connectionTime}$' is used. This is simply a threshold cap on the amount of new data the robot can receive, and in simulation, the lower the value the more often the robot will go back and transfer. 

\subsubsection{Intermittent Rendezvous Communication System}
The second system that will be tested is a rendezvous system between robots, a decentralized system. To represent realistic communication constraints, each robot will have two communication systems: one with low bandwidth but high range, and another with high bandwidth but limited range. When a robot decides to transfer data it will send a rendezvous request to another robot using its high-range system. Other robots will respond and use this communication mode to discuss a meetup location, and once agreed both robots will travel to this location. While traveling the robots will not do any exploration and only choose known and safe paths. When both are at this location they will transfer all the data they have with each other using this high-bandwidth system. Similar to the fixed base station system, the robot will decide to rendezvous when hitting a threshold called `$\mathrm{connectionTime}$'.

\begin{algorithm}
\caption{Robot Behaviors in Mars Exploration Simulation}
\begin{small}
\begin{algorithmic}[1]

\Procedure{Explore}{}
    \State Identify unknown cells adjacent to known cells
    \State Choose a random unknown cell to move to
    \State Generate path to chosen cell using A* algorithm
    \State \textbf{while} moving \textbf{do}
    \State \quad "See" and collect data from cells within a radius
\EndProcedure

\Procedure{CommunicateWithBase}{}
    \If{$connectionTime$ threshold hit}
        \State Calculate most efficient route back to base
        \State Move to base and perform bi-directional data transfer
    \EndIf
\EndProcedure

\Procedure{InitiateRendezvous}{}
    \If{$connectionTime$ threshold hit}
        \State Send rendezvous request to another robot
        \State Agree on meetup location and move to it
        \State Perform data transfer upon meeting
    \EndIf
\EndProcedure

\end{algorithmic}
\end{small}
\end{algorithm}

\subsection{Electrostatic Anomaly Characterization}
We will use the same simulator framework to assess the capabilities of a multi-robot system for locating and estimating the electrostatic potential of Martian dust devils. However as the distances between robots are a lot closer, they do not use the proposed communication system described earlier and we assume they have close range communication for our initial analysis. 
\subsubsection{Environment Description}
Instead of using a varying terrain map, we assume a flat terrain with no hazardous zones, to remove any kind of limits on the movement behavior of the robots, which might affect baseline results. The size of the map is 100$\times$100, where one cell is equal to one meter. This scale was chosen to make feasible the detection of dust devils of a similar size to Earth-bound measurements of dust devils. As there have not been any electrostatic measurements of dust devils on Mars, we use representative numbers from dust devils on Earth \cite{crozier1970dust}. To generate the electrostatic fields on this grid world, we use the inverse square law $\frac{1}{r^2}$ to determine the field propagation in the grid world. 

\begin{figure}[htp]
    \centering
    \includegraphics[width=0.3\textwidth]{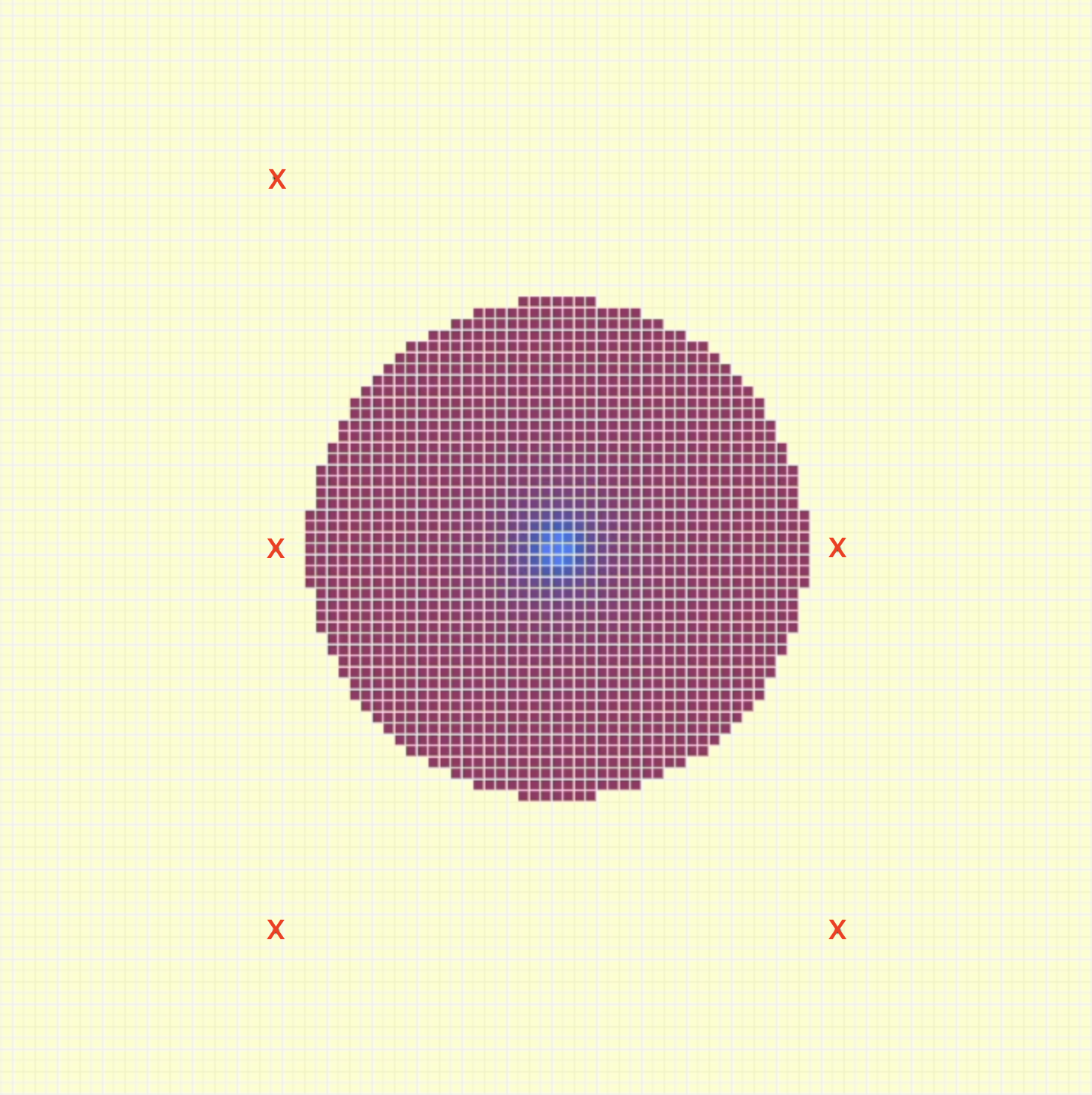}
    \caption{An example of initial conditions for an electrostatic anomaly experiment. The colour of the grid represents the electrostatic environment, and red crosses the robots.}
    \label{fig:ExampleSetup}
\end{figure}

To evaluate how well the system works, we will assume that all robots are within communication range and will not need to rendezvous with each other as in Section IIIA. The simulation will be run multiple times whereby the dust devil is placed in a random position in the grid. Robots will also be placed at fixed, equidistant points across the map, dependent on how many robots are used: Fig.~\ref{fig:ExampleSetup} shows an example of the simulation environment. During the simulation, each robot uses a predictive algorithm called `$\mathrm{extended\_gradients}$'  to determine the location and strength of the dust devil. Two different setups are evaluated: one where electrostatic measurements are shared between robots and one where data is not shared. We assume that sharing data will allow for greater accuracy of the predictions as the robots will have different points of view of the dust devil. However, this could come at a cost of longer processing time and potential error-prone predictions. 
To accurately assess the performance of these different configurations, we run each simulation 20 times with both two robots and five robots, placing the dust devil in a random location for each run. Our evaluation will focus on two key metrics: the error between the predicted and actual locations (distance error) and the precision of the strength estimation (strength error). This will allow us to identify the relative performance of the two systems behave and whether it might be feasible to use a multi-robot system for this application.

\subsubsection{Dust Devil Locating Algorithm}
The algorithm that will be used to predict both the location and the strength of a dust devil is called `$\mathrm{extended\_gradients}$'. As the robot travels through the grid world, it records the electrostatic field at each cell it visits. This data can then be used to create a gradient field, which is represented as a vector at each cell, with a direction and magnitude where a smaller gradient signifies a small change in the electric field. As the robot visits more cells it gains more electrostatic gradient information. It uses the gradient vector direction component to decide its next movement. To ensure there is a mix of exploration of the grid world, each robot has a strategy where there is a 20\% chance of moving randomly or moving towards the gradient direction with a 45-degree deviation, which better allows for the identification of novel gradient information.

\begin{figure}[t!]
    \centering
    \includegraphics[width=0.42\textwidth]{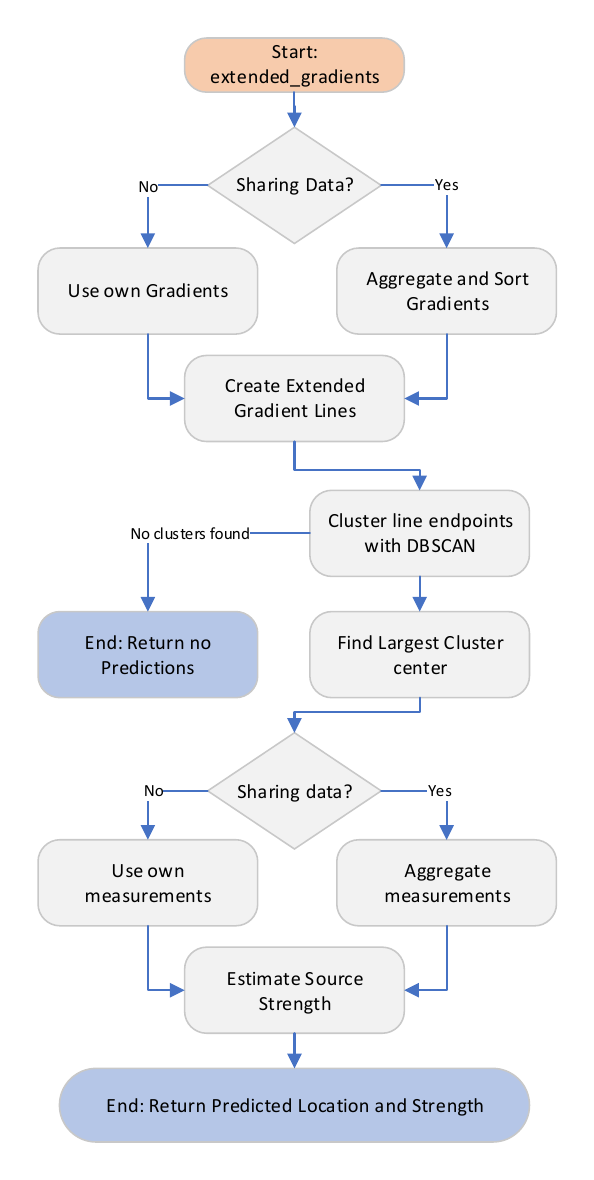}
    \caption{Flowchart of the `$\mathrm{extended\_gradients}$' algorithm. }
    \label{fig:flowchart}
\end{figure}

At each time step the robot uses this gradient field to predict the location and the strength of the dust devil. The full algorithm can be seen explained in the flowchart shown in Fig. \ref{fig:flowchart}. There are four main parts (Algorithms 2--5):
\begin{itemize}
    \item \verb|Aggregate and Sort Gradients|: This part of the algorithm is only run when the robots are sharing data. It collects the gradients from all other robots in the simulation and filters them with the top 5 strongest magnitudes, this then gets added to the focal robot's gradients. 
    \item \verb|Create Extended Gradient Lines|: This part of the algorithm takes all the gradients collected and firstly, normalises them and then, extends them by a scaling factor and the reverse of the inverse square law. This creates lines from the cell location to the predicted endpoint. 
    \item \verb|Cluster line endpoints with DBSCAN|: This uses a clustering algorithm called DBSCAN \cite{khan2014dbscan} to cluster all line endpoints together . 
    \item \verb|Estimate Source Strength|: Here we take the top 20 most recent electrostatic measurements (either from one robot or the shared measurements) and calculate the distance between the location of the point readings and the predicted location. This is then used with the inverse square law to give an estimate of the dust devil's strength. Some filtering is done by using electrostatic measurements that have low magnitude as this means the measurement is far away from the source allowing for some error in the predictions. 
\end{itemize}

Pseudo-code of each of these parts can be found in Section \ref{section:psudocode}

\label{section:psudocode}
\begin{algorithm}
\caption{Aggregate and Sort Gradients}
\begin{small}
\begin{algorithmic}[1]
\State Initialize $aggregatedInfo$ with own gradients
\State Initialize $temp$ as an empty list

\For{each robot in the model excluding self}
    \State Add robot's gradients to $temp$
\EndFor

\State Sort $temp$ by the last element in descending order of magnitude 
\State $N \gets 5$ \Comment{Top items to select}
\State Select top $N$ items from $temp$ into $topNAggregatedInfo$
\State Extend $aggregatedInfo$ with $topNAggregatedInfo$
\end{algorithmic}
\end{small}
\end{algorithm}

\begin{algorithm}
\caption{Create Extended Gradient Lines}
\begin{small}
\begin{algorithmic}[1]
\State Initialize $lines$ as an empty list
\State $scalingFactor \gets 5$

\For{each $(x, y, dx, dy, magnitude)$ in $aggregatedInfo$}
    \State $dx \gets -dx$
    \State $dy \gets -dy$
    \State $gradNorm \gets \sqrt{dx^2 + dy^2}$
    
    \If{$gradNorm \neq 0$}
        \State Normalize $(dx, dy)$ by $gradNorm$
        \State $lineLength \gets scalingFactor \cdot \sqrt{\frac{1}{magnitude + 1e-6}}$
        \State $endX \gets x + dx \cdot lineLength$
        \State $endY \gets y + dy \cdot lineLength$
        \State Append $((x, y), (endX, endY))$ to $lines$
    \EndIf
\EndFor
\end{algorithmic}
\end{small}
\end{algorithm}
\vspace{-5mm}

\begin{algorithm}
\caption{Cluster line endpoints with DBSCAN}
\begin{small}
\begin{algorithmic}[1]
\State $endPoints \gets$ Extract end points from $lines$
\If{$length(endPoints) > 0$}
    \State Perform DBSCAN clustering on $endPoints$
    \State Calculate $clusterCenters$ as the mean of points in each cluster
\EndIf

\end{algorithmic}
\end{small}
\end{algorithm}

\begin{algorithm}
\caption{Estimate Source Strength}
\begin{small}
\begin{algorithmic}[1]
\State Convert $predictedSource$ to tuple
\State Extract and sort top 20 recent measurements by step
\State Initialize $sourceStrengthEstimates$ as empty
\For{each measurement in sorted recent measurements}
    \State Calculate distance $r$ and strength $S$ from $predictedSource$
    \If{$r > 0$}
        \State Add $S$ to $sourceStrengthEstimates$
    \EndIf
\EndFor
\State $trimFraction \gets 0.2$
\If{not empty $sourceStrengthEstimates$}
    \State $finalEstimate \gets$ Trimmed mean of $sourceStrengthEstimates$
\Else
    \State $finalEstimate \gets$ None
\EndIf
\end{algorithmic}
\end{small}
\end{algorithm}
\begin{figure*}[t!]
    \centering
    \includegraphics[width=0.85\textwidth]{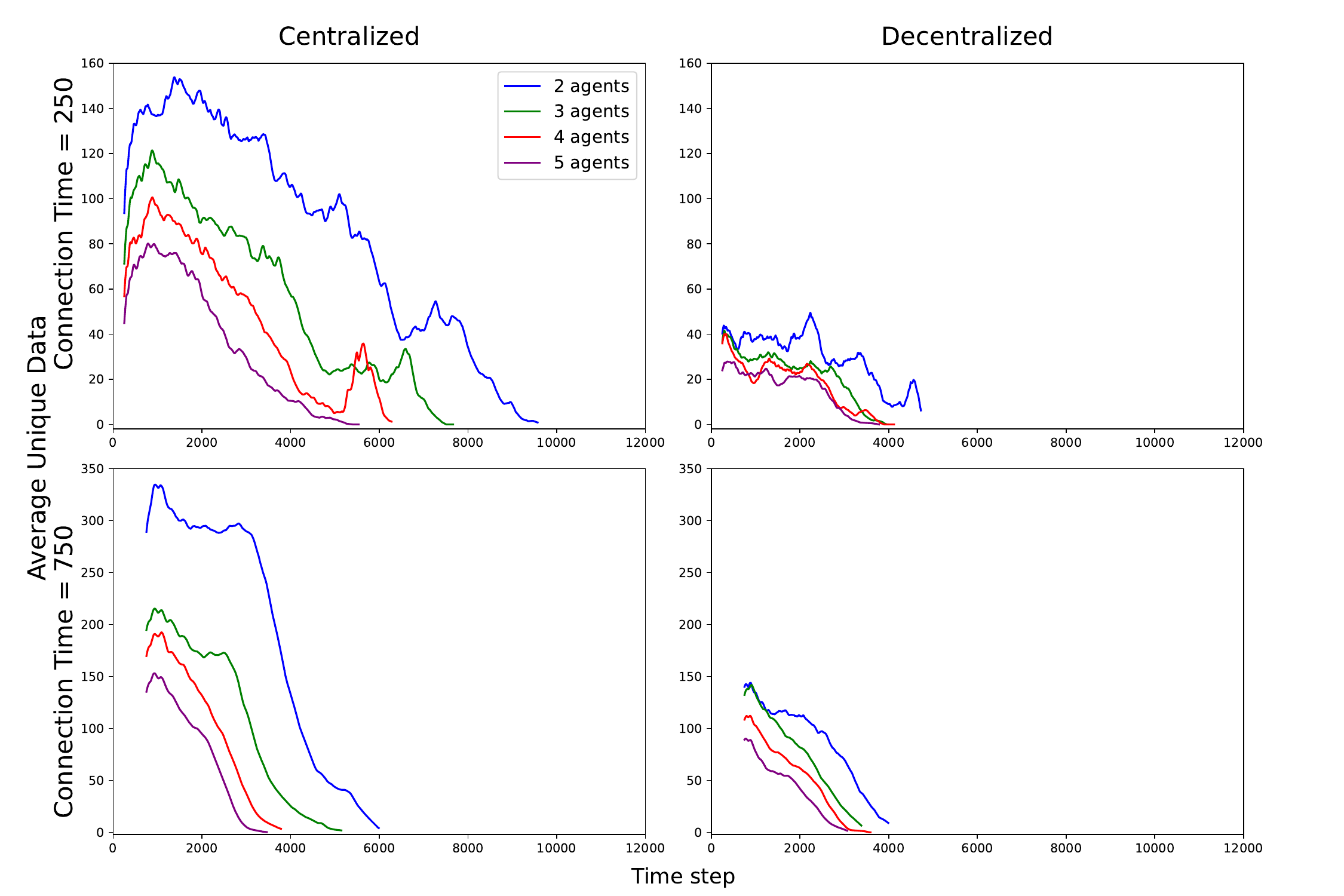}
    \caption{A comparison of the communication-constrained exploration strategies: average unique data over time on flat terrain map with minimum and maximum \textit{connectionTime} settings. The exploration time is longer and the unique (most at-risk) data is higher for the centralized (base station) communication strategy.}
    \label{fig:noTerrainMap}
\end{figure*}

\begin{figure*}[t!]
    \centering
    \includegraphics[width=0.85\textwidth]{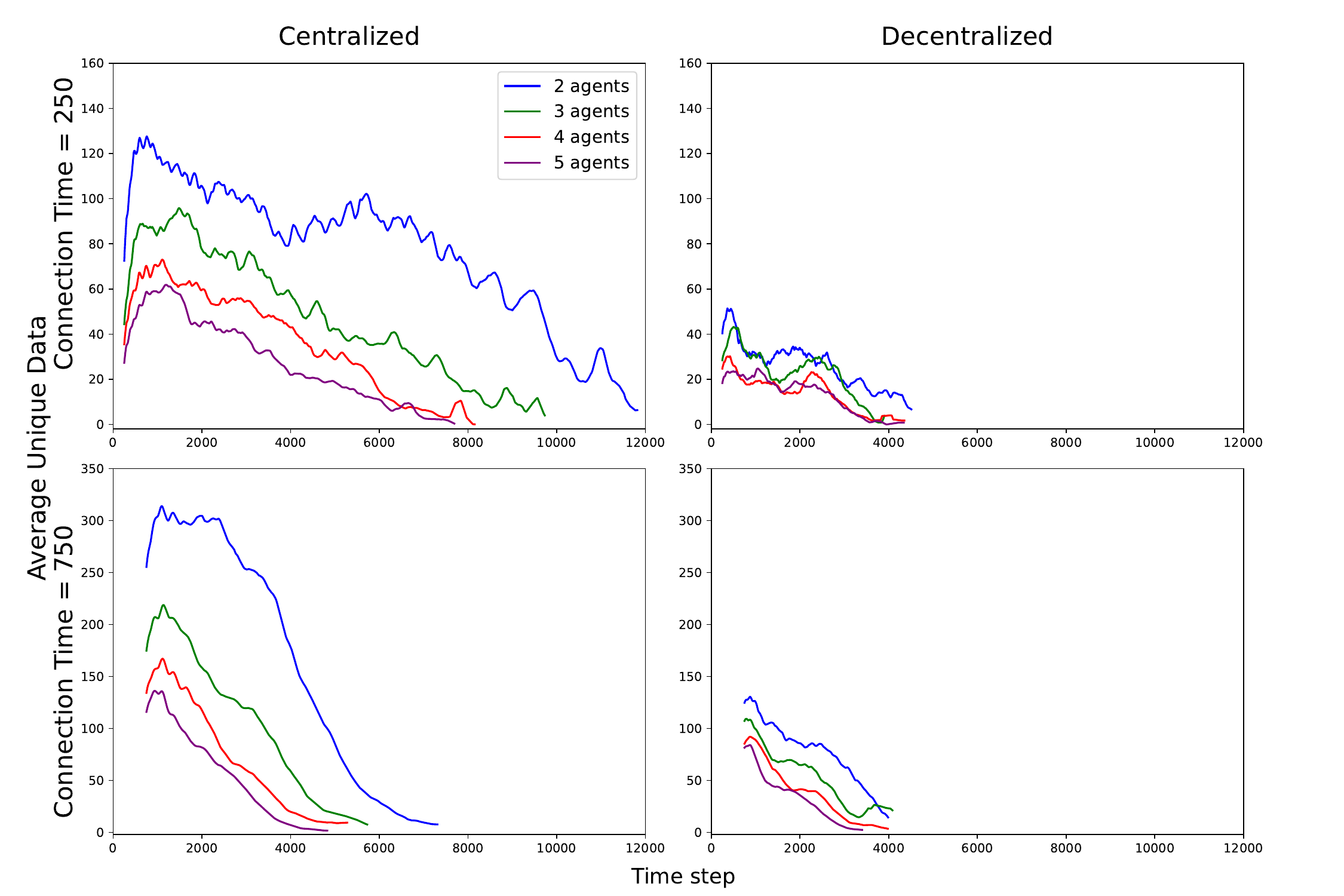}
    \caption{Average unique data on rugged terrain map with minimum and maximum \textit{connectionTime} settings. Results are similar overall as for the flat terrain map, with a slightly slower traversal, less evident in the rendezvous strategy. There is slightly less data collected because parts of the map are inaccessible (83\% is accessible).}
    \label{fig:TerrainMap}
\end{figure*}

\section{Results}
\subsection{Communication-Constrained Exploration}
In Figs.~\ref{fig:noTerrainMap} and \ref{fig:TerrainMap}, the final results of the simulation trials can be seen. Each plot shows the different maps used, one with rugged terrain and one with flat terrain, where the top row shows the centralized and decentralized systems with a low $\mathrm{connectionTime}$, and the bottom row shows it with a high $\mathrm{connectionTime}$. The y-axis shows the average amount of unique data each robot has, whereby the higher the value the more risk there is and more data would be removed if a robot is lost. The $x$-axis shows the amount of steps or time it took to fully explore the area. Note that in the rugged terrain map, approximately 85\% of the map can be explored as some cells are too hazardous for the robots to visit.

From the plots presented, we can get some interesting insights. When looking at the results from the flat terrain map, Fig.~\ref{fig:noTerrainMap}, shows that the fixed base station system requires a significantly longer time to complete the map exploration when only two robots are present. When we add more robots there is a decrease in the time to explore. If we look at the unique data it is quite high, but also decreases with more robots. If we compare this with the rendezvous system, it demonstrates a very short exploration time. It also has a considerably lower amount of unique data, suggesting a lower risk of data loss from prior exploration. 

When we consider the $\mathrm{connectionTime}$ variable, with longer times the fixed base station shows a shorter exploration time, however still not as much as using the rendezvous system. When we switch and look at the rugged terrain map in Fig.~\ref{fig:TerrainMap}, the same results can be seen, indicating that the terrain does not influence the exploration time and unique data when comparing the two systems.

The plots for several results show a plateau, or in some cases, even an increase in unique data. This happens around the midpoint of the exploration. This may result from a stage during exploration where the robots encounter many more new cells. As the exploration goes on and fewer unvisited cells are left, this plateau decreases again. 
 
One of the reasons for the long exploration time in the fixed base station system with low connection time is that the robots are spending a long time going back and forth to the base station and less time exploring the area. This is not the case in the rendezvous system as the travel time spent going to meet another robot is a lot less as the two robots meeting up will find a midpoint between them.  The efficiency of the robots during exploration was assessed over multiple trials: this was done by comparing the ratio of movements made to the number of environment cells discovered. The results of these trials can be seen in Table \ref{table:results}. It shows that the pairwise communication strategy is more efficient than the base station.

\begin{table}[h]
\resizebox{\columnwidth}{!}{%
\begin{tabular}{llllll}
\textbf{Pairwise   communication} & 1 Agent & 2 Agents & 3 Agents & 4 Agents & 5 Agents \\
Movement/ Cell seen average & 0.25 & 0.22 & 0.19 & 0.20 & 0.18 \\
Movement/ Cell seen std\_dev & 0.012 & 0.008 & 0.018 & 0.020 & 0.014 \\
 &  &  &  &  &  \\
\textbf{Base station   communication} & 1 Agent & 2 Agents & 3 Agents & 4 Agents & 5 Agents \\
Movement/ Cell seen average & 0.67 & 0.40 & 0.36 & 0.29 & 0.27 \\
Movement/ Cell seen std\_dev & 0.015 & 0.029 & 0.018 & 0.027 & 0.030
\end{tabular}%
}
\caption{Exploration efficiency results of pairwise and base station communication strategies.}
\label{table:results}
\end{table}

\subsection{Electrostatic Anomaly Characterization}
\begin{figure*}[t!]
\centering
\begin{subfigure}{0.3\textwidth}
    \centering
    \includegraphics[width=\textwidth]{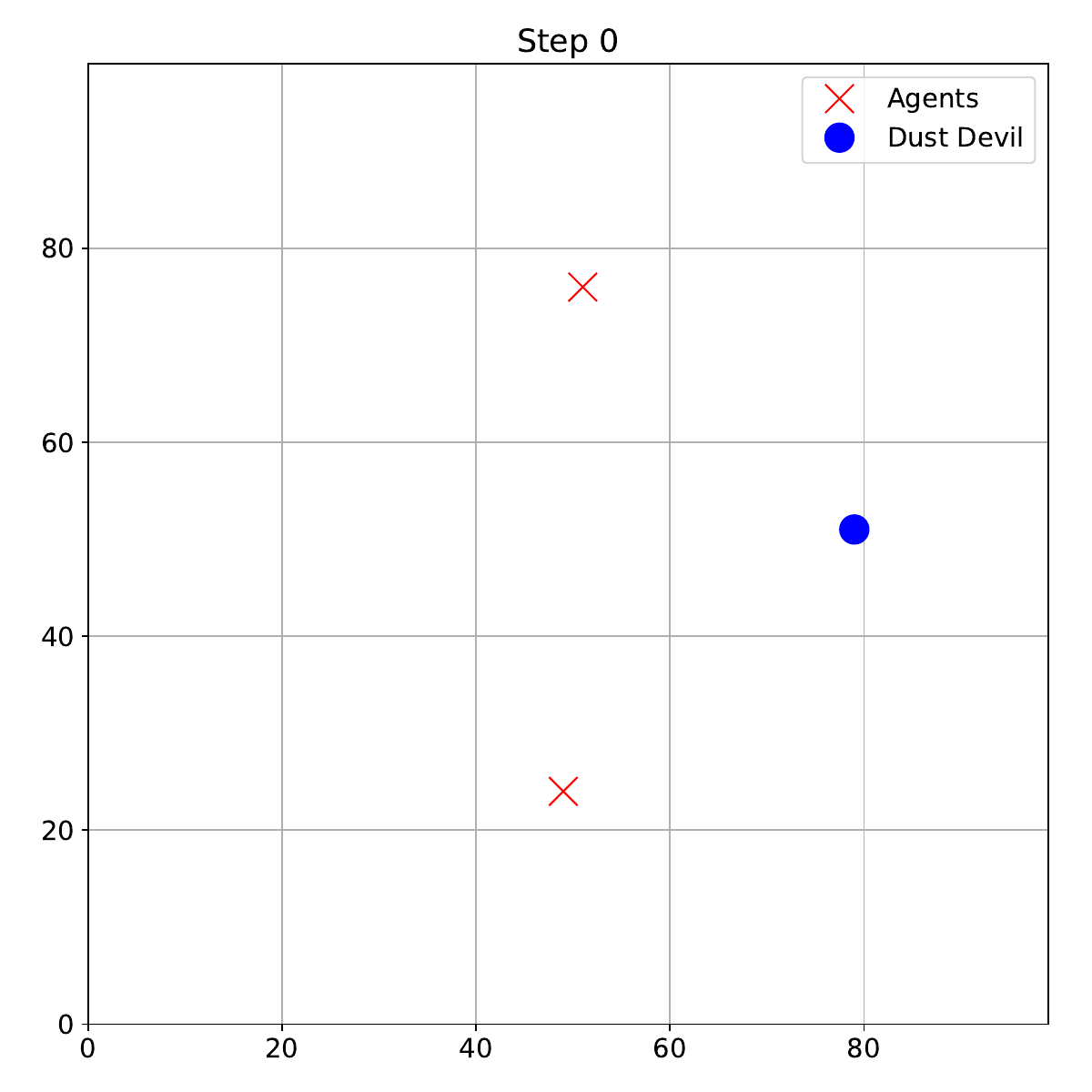}
    \caption{Configuration at time step 0}
    \label{fig:single0Step}
\end{subfigure}%
\begin{subfigure}{0.3\textwidth}
    \centering
    \includegraphics[width=\textwidth]{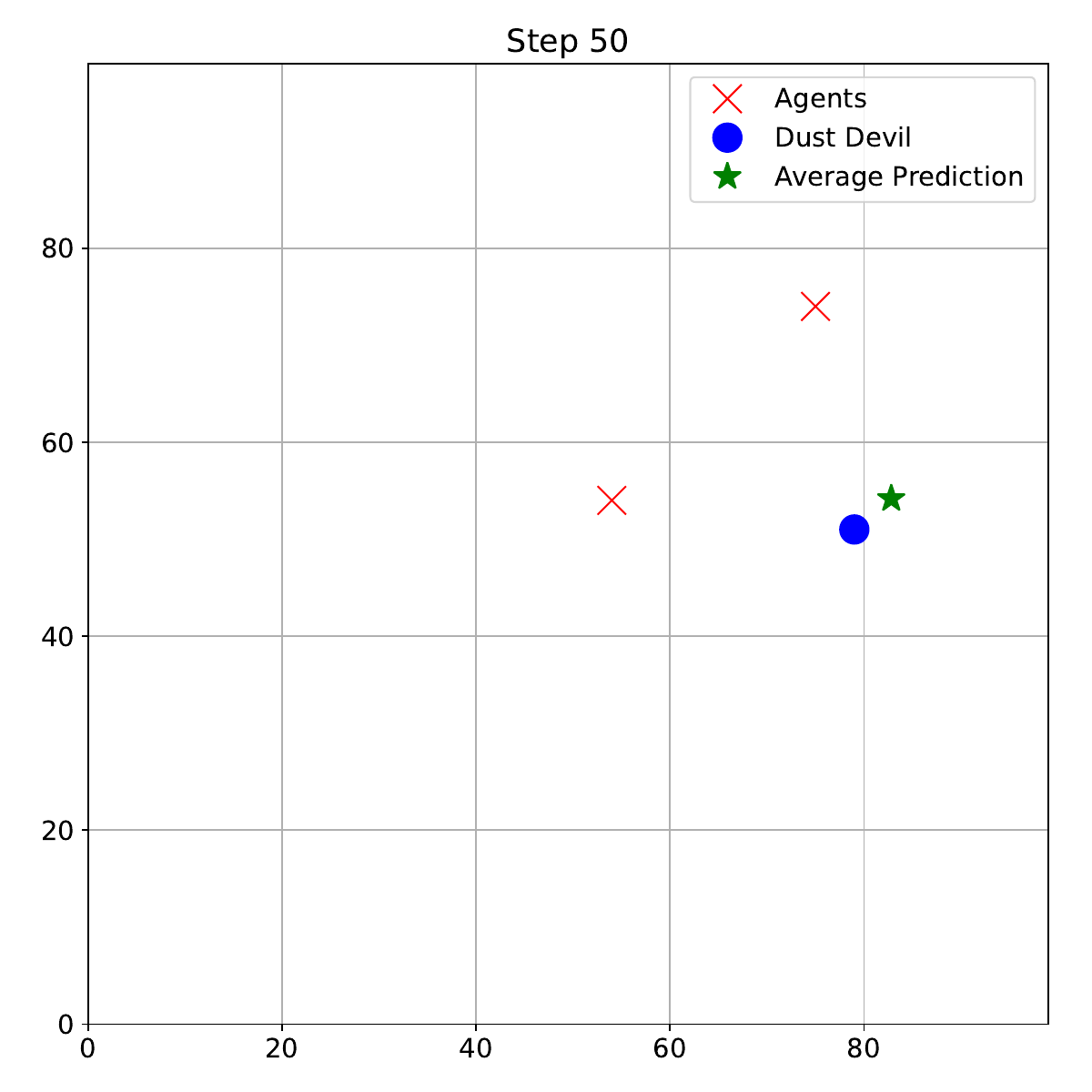}
    \caption{Time step 50}
    \label{fig:single50Step}
\end{subfigure}%
\begin{subfigure}{0.3\textwidth}
    \centering
    \includegraphics[width=\textwidth]{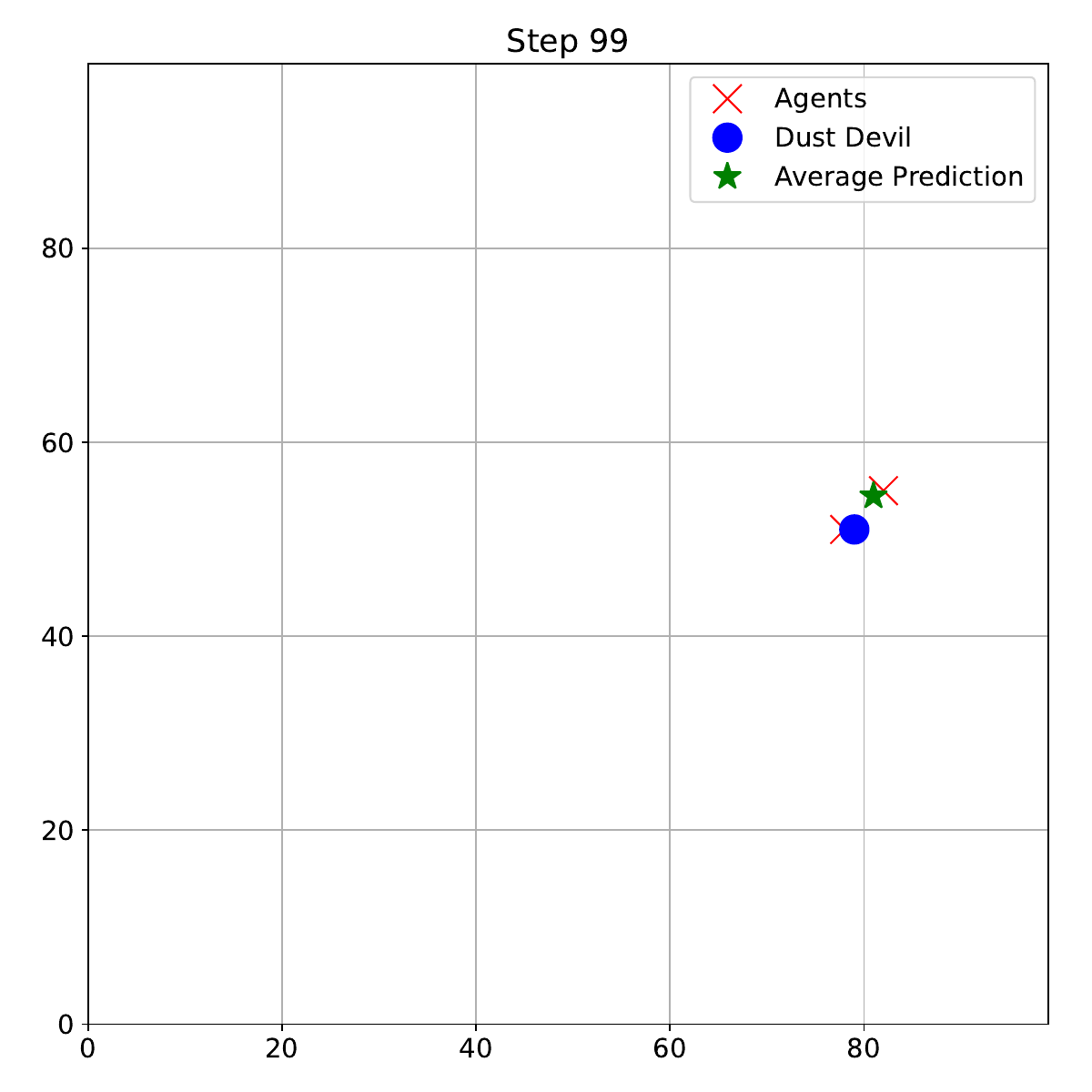}
    \caption{Time step 99}
    \label{fig:single99Step}
\end{subfigure}

\caption{Single simulation run of `$\mathrm{extended\_gradients}$' algorithm over time (2 robots).}
\end{figure*}

\begin{figure*}[t!]
\centering
\begin{subfigure}{0.475\textwidth}
    \includegraphics[width=1\textwidth]{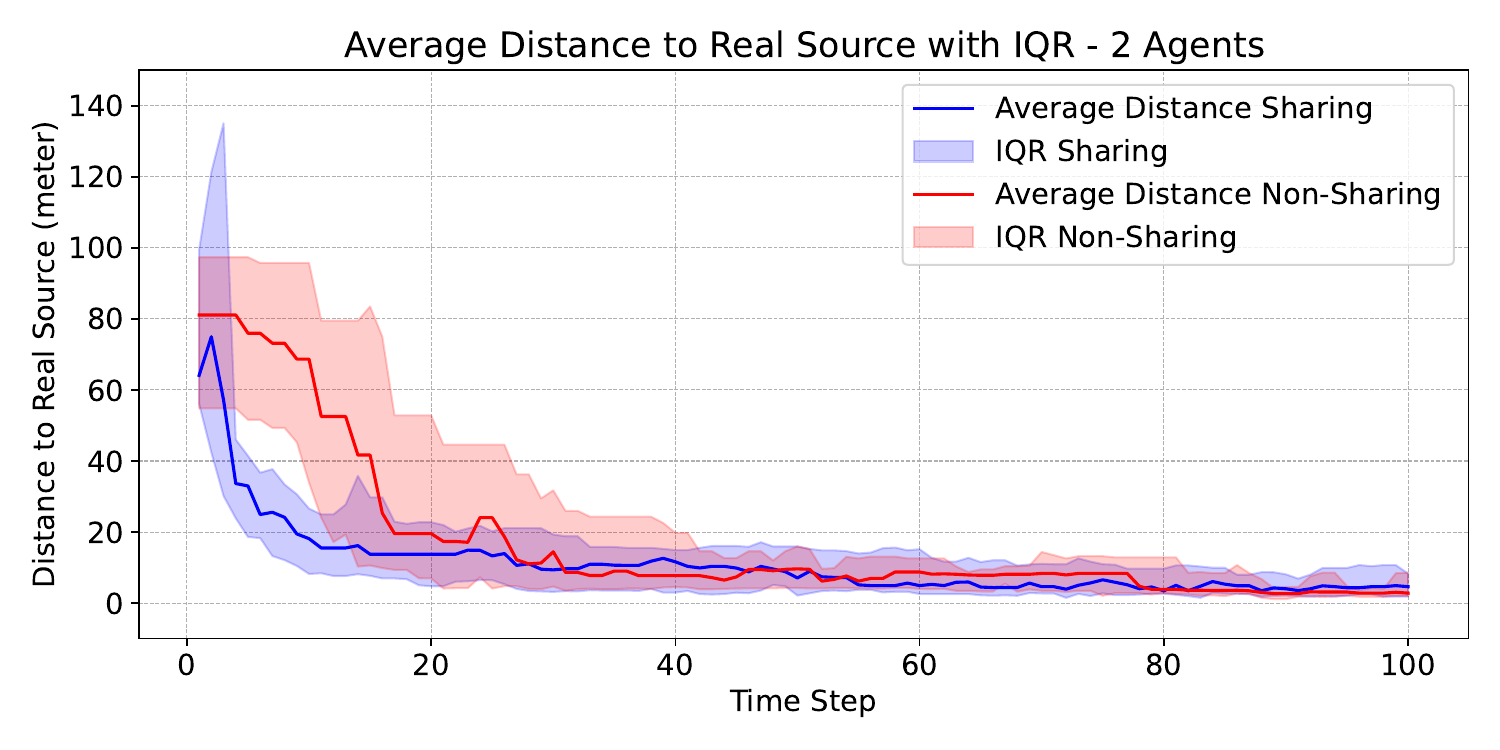}
    \caption{20 runs showing distance error of prediction with 2 robots.}
    \label{fig:2Agents20Runs}
\end{subfigure}
\hfill
\begin{subfigure}{0.475\textwidth}
    \includegraphics[width=1\textwidth]{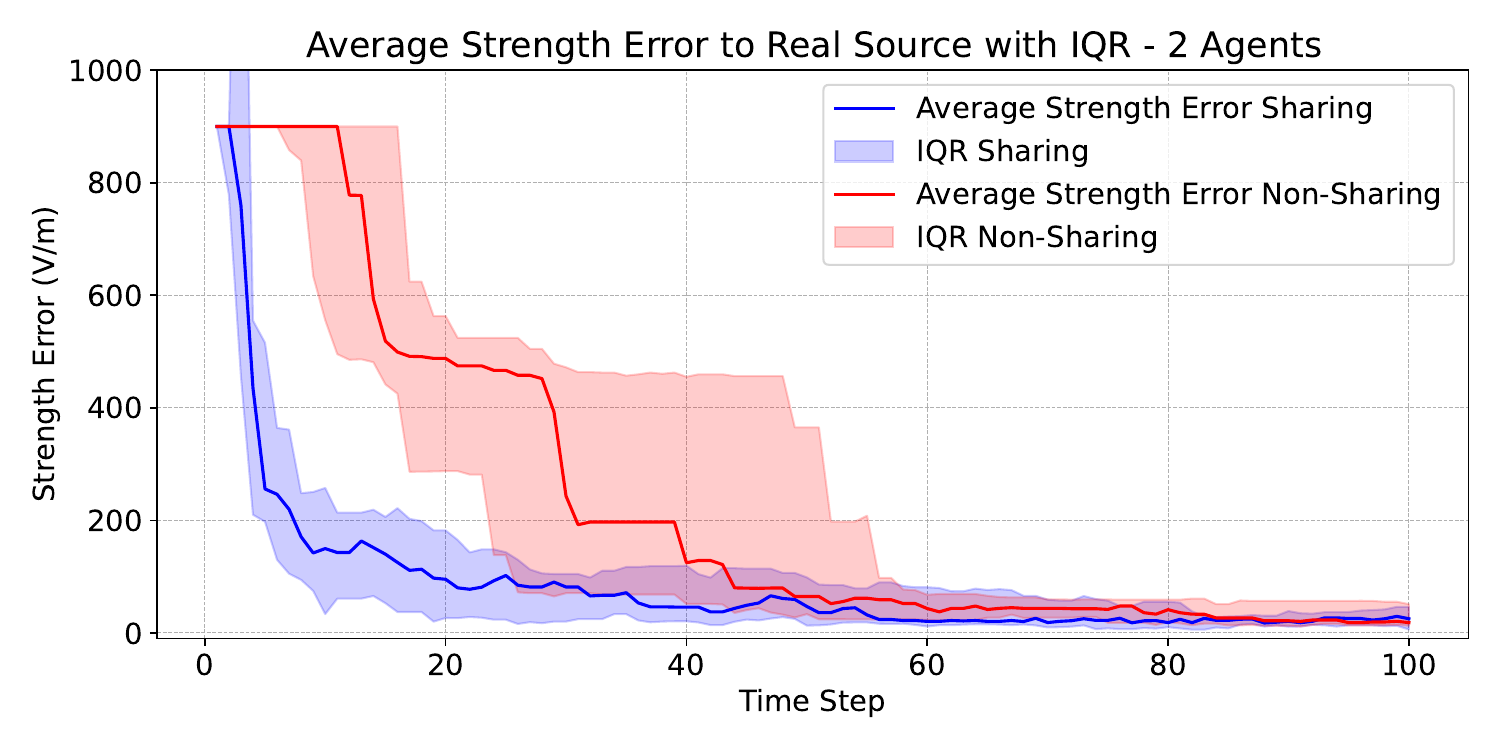}
    \caption{20 runs showing strength error of prediction with 2 robots.}
    \label{fig:2Agents20RunsStrength}
\end{subfigure}

\vskip\baselineskip

\begin{subfigure}{0.475\textwidth}
    \includegraphics[width=1\textwidth]{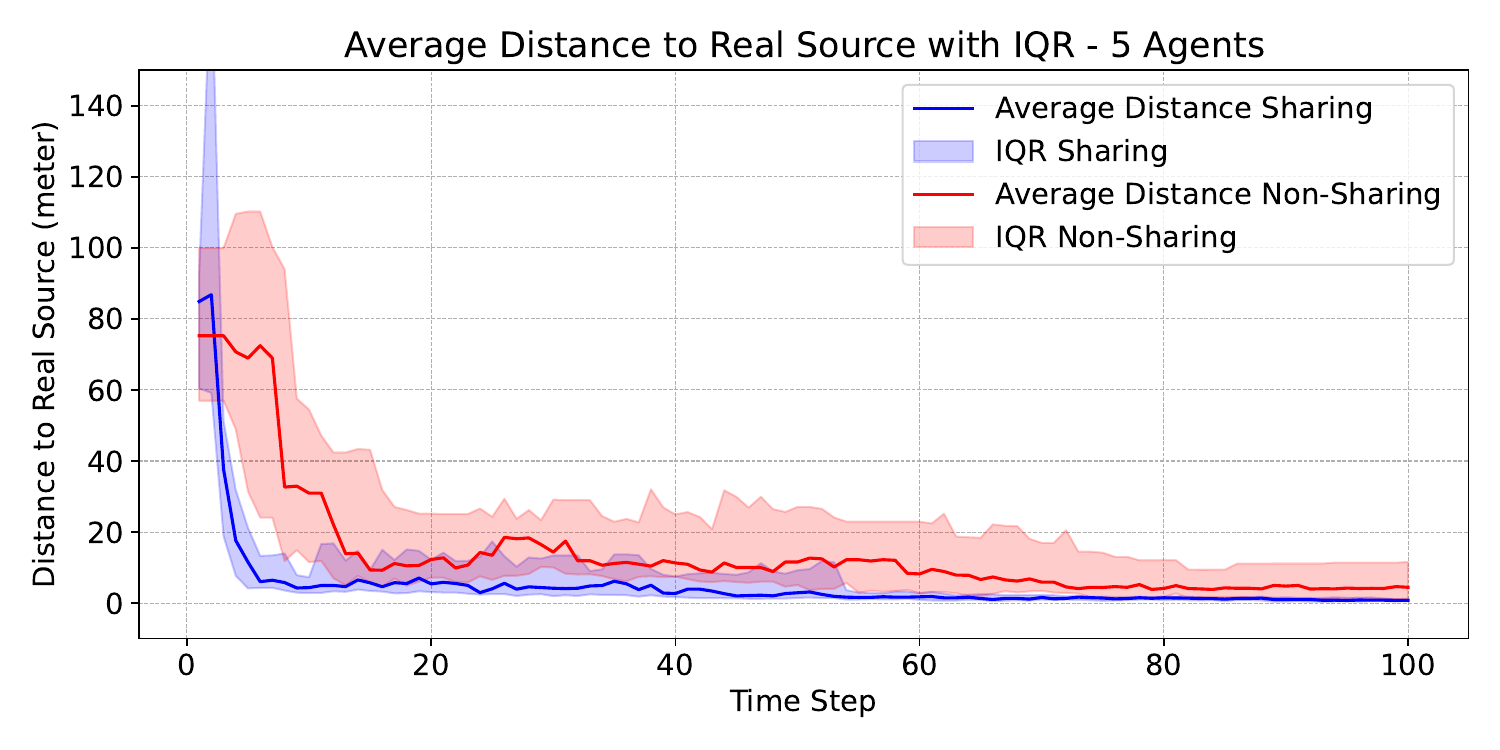}
    \caption{20 runs showing distance error of prediction with 5 robots.}
    \label{fig:5Agents20Runs}
\end{subfigure}
\hfill
\begin{subfigure}{0.475\textwidth}
    \includegraphics[width=1\textwidth]{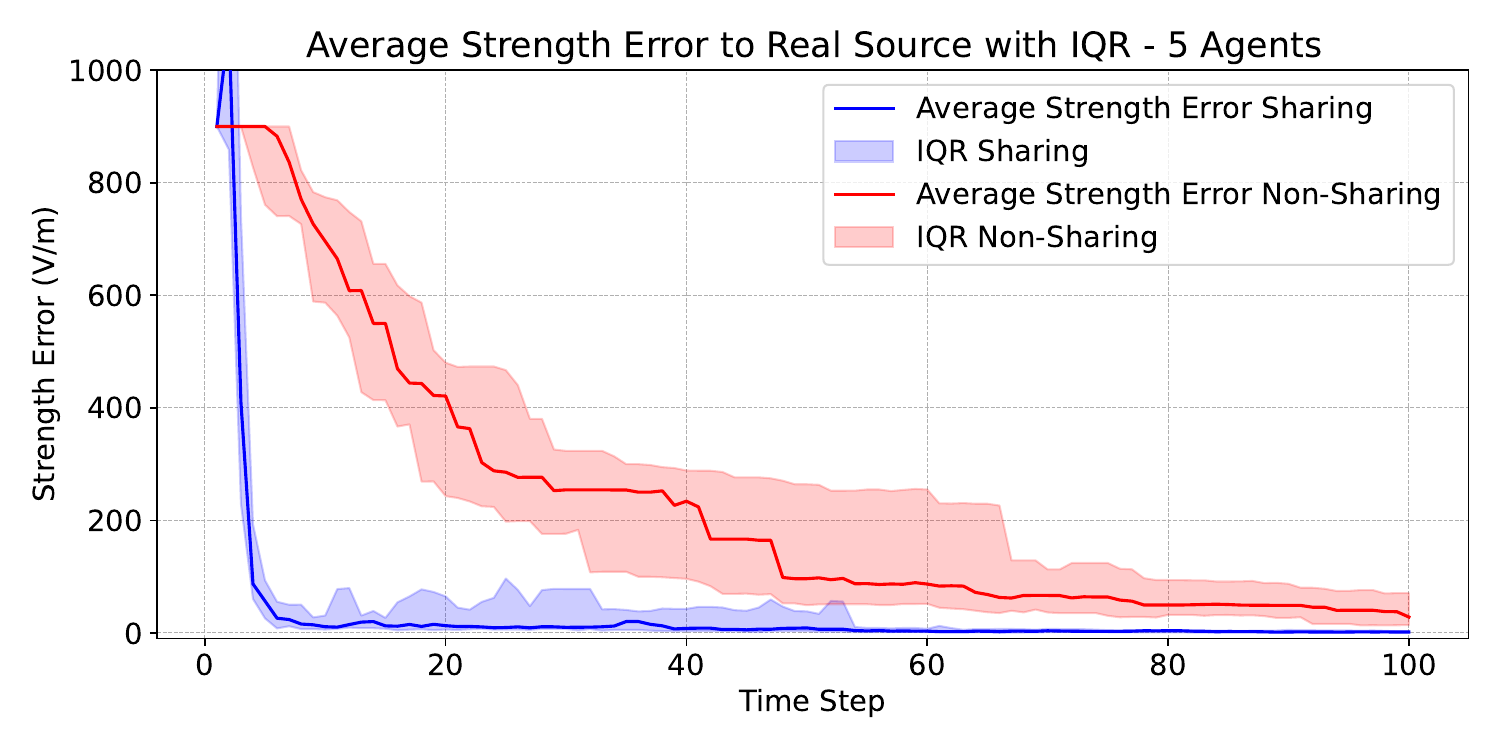}
    \caption{20 runs showing strength error of prediction with 5 robots.}
    \label{fig:5Agents20RunsStrength}
\end{subfigure}
\caption{Plots of multiple simulation runs with both 2 and 5 robots. Error bands show interquartile range.}
\end{figure*}

In Figs. \ref{fig:single0Step}, \ref{fig:single50Step} and \ref{fig:single99Step} a single run can be seen for both two robots over time. We have put a cap on the step count to 100 as anything after this would mean the robot will have more than enough time to traverse the grid world and move on top of the dust devil. The plots show the movements of the robots over time with the average predictions that they make. 

In Figure \ref{fig:2Agents20Runs},\ref{fig:2Agents20RunsStrength}, \ref{fig:5Agents20Runs}, and \ref{fig:5Agents20RunsStrength} a full set of simulation runs can be seen. These are done over 20 runs with both 2 robots and 5 robots. Figures \ref{fig:2Agents20Runs} and \ref{fig:5Agents20Runs} show the distance error, while Figures \ref{fig:2Agents20RunsStrength} and \ref{fig:5Agents20RunsStrength} show the strength error. In all cases, the blue lines show simulations where data is shared between robots and the red where robots do not use any shared information. All of these graphs show the interquartile range as the error bands.

\section{Discussion and Conclusion}
\subsection{Communication-Constrained Exploration}
We considered two communication strategies for a multi-robot system in a remote environment without preexisting communications infrastructure. We particularly consider the multi-robot SLAM case where large amounts of terrain data might need to be shared to build a shared map of an unfamiliar environment as it is explored. If robots have to be in physical proximity to share data, our two strategies involving a fixed base station (centralized) and a pairwise rendezvous system (decentralized) would both be relevant. We found that the time for exploration for the rendezvous system consistently outperforms the fixed base station system, regardless of the number of robots or the structure of the terrain. Looking at the amount of unique data (located in only a single robot) in each system it is also lower in the rendezvous system suggesting it has lower risk of costly data loss on robot failure. 

Clearly, the choice of communication strategies plays a guiding role in the efficiency of multi-robot exploration. Nevertheless, a decentralized swarm SLAM approach has its challenges \cite{Kegeleirs2021}, and in some cases, there may be good reasons to maintain a central base station with particular communication capabilities: to a relay satellite, for example. In that case, an approach combining elements of both centralization and decentralization could be an attractive choice. Future research could test the rendezvous strategy with real robots, for example coordinating meetings via LoRa and then sharing data via WiFi. We will also employ realistic multi-robot simulations to develop our communication strategies to better consider the impact of rugged terrain on communication performance.

\subsection{Electrostatic Anomaly Characterization}
Our investigation into dust devil location and strength prediction has provided insights into the capabilities and limitations of using a multi-robot system for this task. Using the simulation framework, we explored various scenarios involving different dust devil locations with different numbers of robots and whether they were sharing information or not. The core of our experiments was the `$\mathrm{extended\_gradients}$' algorithm, which is designed to predict the location and strength of dust devils based on electrostatic measurements. A notable finding from our study is the impact of information sharing between robots: this improved the speed and accuracy of finding the location and strength of the electrostatic anomaly. However, in the early parts of the simulation, shared information led to a larger error in predicting the location of the source, which in turn also created a large error in the strength prediction. Having done simulations for both a 2-robot setup and a 5-robot setup, we found larger group sizes extended the benefit of a faster prediction, but also a larger inaccuracy at the start of the simulation. This means there needs to be a balance between the speed of locating the source while also managing the errors at the start. This initial error highlights the challenges of using multi-robot systems for this kind of task, and more research needs to be done to refine the algorithm. 

\subsection{Conclusion and future works}
This research shows the importance of selecting an appropriate communication strategy in multi-robot systems for efficient exploration, whereby the decentralized, pairwise rendezvous system, shows particular promise due to its superior performance and lower risk profile. However, the use of the centralized system could offer a balanced solution, in unique scenarios requiring a specific communication capability. For electrostatic anomaly characterization, the `$\mathrm{extended\_gradients}$' algorithm offers a promising approach. We will investigate cases where an area contains more than one anomaly, which is conceivable given an estimate of one dust devil per square km per sol \cite{jackson2018framework}. While the approach works well with a static (or possibly, slow-moving) dust devil, real-world data shows that dust devils on Earth often move at significant speed (5 to 10 ms$^{-1}$, \cite{crozier1970dust}). The algorithm will also need development for such cases.

\bibliographystyle{IEEEtran}
\bibliography{references}

\begin{thebibliography}{10}
\providecommand{\url}[1]{#1}
\csname url@rmstyle\endcsname
\providecommand{\newblock}{\relax}
\providecommand{\bibinfo}[2]{#2}
\providecommand\BIBentrySTDinterwordspacing{\spaceskip=0pt\relax}
\providecommand\BIBentryALTinterwordstretchfactor{4}
\providecommand\BIBentryALTinterwordspacing{\spaceskip=\fontdimen2\font plus
\BIBentryALTinterwordstretchfactor\fontdimen3\font minus \fontdimen4\font\relax}
\providecommand\BIBforeignlanguage[2]{{%
\expandafter\ifx\csname l@#1\endcsname\relax
\typeout{** WARNING: IEEEtran.bst: No hyphenation pattern has been}%
\typeout{** loaded for the language `#1'. Using the pattern for}%
\typeout{** the default language instead.}%
\else
\language=\csname l@#1\endcsname
\fi
#2}}

\bibitem{ruf2009emission}
C.~Ruf, N.~O. Renno, J.~F. Kok, E.~Bandelier, M.~J. Sander, S.~Gross, L.~Skjerve, and B.~Cantor, ``Emission of non-thermal microwave radiation by a {Martian} dust storm,'' \emph{Geophysical Research Letters}, vol.~36, no.~13, 2009.

\bibitem{Balaram2021}
J.~Balaram, M.~M. Aung, and M.~P. Golombek, ``{The Ingenuity Helicopter on the Perseverance Rover},'' \emph{Space Science Reviews}, vol. 217:56, 2021.

\bibitem{Chahat2020helicopter}
N.~Chahat, J.~Miller, E.~Decrossas, L.~McNally, M.~Chase, C.~Jin, and C.~Duncan, ``{The Mars Helicopter Telecommunication Link: Antennas, Propagation, and Link Analysis},'' \emph{IEEE Antennas and Propagation Magazine}, vol.~62, no.~6, pp. 12--22, 2020.

\bibitem{Laine2018}
M.~Laîné, C.~Tamakoshi, M.~Touboulic, J.~Walker, and K.~Yoshida, ``Initial design characteristics, testing and performance optimisation for a lunar exploration micro-rover prototype,'' \emph{Advances in Astronautics Science and Technology}, vol.~1, pp. 111--117, 2018.

\bibitem{Kolvenbach2019}
H.~Kolvenbach, E.~Hampp, P.~Barton, R.~Zenkl, and M.~Hutter, ``Towards jumping locomotion for quadruped robots on the moon,'' in \emph{2019 IEEE/RSJ International Conference on Intelligent Robots and Systems (IROS)}, 2019, pp. 5459--5466.

\bibitem{Uno2021}
K.~Uno, N.~Takada, T.~Okawara, K.~Haji, A.~Candalot, W.~F.~R. Ribeiro, K.~Nagaoka, and K.~Yoshida, ``Hubrobo: A lightweight multi-limbed climbing robot for exploration in challenging terrain,'' in \emph{2020 IEEE-RAS 20th International Conference on Humanoid Robots (Humanoids)}, 2021, pp. 209--215.

\bibitem{jaxa}
\BIBentryALTinterwordspacing
{Japan Aerospace Exploration Agency}, ``{JAXA | Smart Lander for Investigating Moon (SLIM)},'' accessed: 2024-02-15. [Online]. Available: \url{https://global.jaxa.jp/projects/sas/slim/}
\BIBentrySTDinterwordspacing

\bibitem{leblanc2008planetary}
F.~Leblanc, K.~Aplin, Y.~Yair, G.~Harrison, J.~P. Lebreton, and M.~Blanc, \emph{Planetary Atmospheric Electricity}.\hskip 1em plus 0.5em minus 0.4em\relax Springer Science \& Business Media, 2008, vol.~30.

\bibitem{zijlstra2024}
G.~Zijlstra, K.~L. Aplin, and E.~R. Hunt, ``Exploring the use of terrestrial robots for atmospheric electricity measurement,'' in \emph{J. Phys. Conf. Ser}, vol. 2702:012006.\hskip 1em plus 0.5em minus 0.4em\relax IOP Publishing, 2024.

\bibitem{dustdevil}
R.~Singh and A.~S. Arya, ``{Martian Dust Devils Observed By Mars Colour Camera Onboard Mars Orbiter},'' \emph{50th Lunar and Planetary Science Conference 2019 (LPI Contrib. No. 2132)}, 2019.

\bibitem{nasadust}
\BIBentryALTinterwordspacing
{NASA Jet Propulsion Laboratory (JPL)}, ``Martian whirlwind takes the `thorofare','' Sept 2023, accessed: 2024-02-08. [Online]. Available: \url{https://www.jpl.nasa.gov/images/pia26074-martian-whirlwind-takes-the-thorofare}
\BIBentrySTDinterwordspacing

\bibitem{crozier1970dust}
W.~Crozier, ``Dust devil properties,'' \emph{Journal of Geophysical Research}, vol.~75, no.~24, pp. 4583--4585, 1970.

\bibitem{zhai2006quasi}
Y.~Zhai, S.~Cummer, and W.~Farrell, ``Quasi-electrostatic field analysis and simulation of {Martian} and terrestrial dust devils,'' \emph{Journal of Geophysical Research: Planets}, vol. 111, no.~E6, 2006.

\bibitem{hornischer2021cimax}
H.~Hornischer, J.~C. Varughese, R.~Thenius, F.~Wotawa, M.~F{\"u}llsack, and T.~Schmickl, ``{CIMAX: collective information maximization in robotic swarms using local communication},'' \emph{Adaptive Behavior}, vol.~29, no.~3, pp. 297--314, 2021.

\bibitem{Valavanis2000}
K.~Valavanis, T.~Hebert, R.~Kolluru, and N.~Tsourveloudis, ``Mobile robot navigation in {2-D} dynamic environments using an electrostatic potential field,'' \emph{IEEE Transactions on Systems, Man, and Cybernetics -- Part A: Systems and Humans}, vol.~30, no.~2, pp. 187--196, 2000.

\bibitem{Pimenta2006}
L.~Pimenta, A.~Fonseca, G.~Pereira, R.~Mesquita, E.~Silva, W.~Caminhas, and M.~Campos, ``Robot navigation based on electrostatic field computation,'' \emph{IEEE Transactions on Magnetics}, vol.~42, no.~4, pp. 1459--1462, 2006.

\bibitem{Bayat2018}
F.~Bayat, S.~Najafinia, and M.~Aliyari, ``Mobile robots path planning: Electrostatic potential field approach,'' \emph{Expert Systems with Applications}, vol. 100, pp. 68--78, 2018.

\bibitem{sancho}
D.~L. Sancho-Pradel and C.~M. Saaj, ``Assessment of artificial potential field methods for navigation of planetary rovers,'' in \emph{2009 European Control Conference (ECC)}, 2009, pp. 3027--3032.

\bibitem{Laine2021}
M.~Laine and K.~Yoshida, ``Multi-rover exploration strategies: Coverage path planning with myopic sensing,'' in \emph{Field and Service Robotics}, G.~Ishigami and K.~Yoshida, Eds.\hskip 1em plus 0.5em minus 0.4em\relax Singapore: Springer Singapore, 2021, pp. 205--218.

\bibitem{Paet2021}
L.~B. Paet, S.~Santra, M.~Laine, and K.~Yoshida, ``Maintaining connectivity in multi-rover networks for lunar exploration missions,'' in \emph{2021 IEEE 17th International Conference on Automation Science and Engineering (CASE)}, 2021, pp. 1539--1546.

\bibitem{Tarapore2020}
D.~Tarapore, R.~Groß, and K.-P. Zauner, ``Sparse robot swarms: Moving swarms to real-world applications,'' \emph{Frontiers in Robotics and AI}, vol.~7, p.~83, 2020.

\bibitem{Kegeleirs2021}
M.~Kegeleirs, G.~Grisetti, and M.~Birattari, ``{Swarm SLAM: Challenges and Perspectives},'' \emph{Frontiers in Robotics and AI}, vol. 8:22969144, 2021.

\bibitem{masad2015mesa}
D.~Masad and J.~Kazil, ``Mesa: an agent-based modeling framework,'' in \emph{Proceedings of the 14th Python in Science Conference (SciPy 2015)}, 2015, pp. 53--60.

\bibitem{khan2014dbscan}
K.~Khan, S.~U. Rehman, K.~Aziz, S.~Fong, and S.~Sarasvady, ``{DBSCAN: Past, present and future},'' in \emph{{The 5th International Conference on the Applications of Digital Information and Web Technologies (ICADIWT 2014)}}.\hskip 1em plus 0.5em minus 0.4em\relax IEEE, 2014, pp. 232--238.

\bibitem{jackson2018framework}
B.~Jackson, R.~Lorenz, and K.~Davis, ``A framework for relating the structures and recovery statistics in pressure time-series surveys for dust devils,'' \emph{Icarus}, vol. 299, pp. 166--174, 2018.

\end{thebibliography}

\end{document}